**Title**

The role of explainability in creating trustworthy artificial intelligence for health care: a comprehensive survey of the terminology, design choices, and evaluation strategies

**Authors**

Aniek F. Markus[a][*] MSc

Jan A. Kors[a] PhD

Peter R. Rijnbeek[a] PhD

[a] Department of Medical Informatics, Erasmus University Medical Center, Rotterdam, The Netherlands

[*] Corresponding author at: Department of Medical Informatics, Erasmus University Medical Center, Rotterdam, The Netherlands. Email address: a.markus@erasmusmc.nl (A.F. Markus)

**Date**

December 3, 2020

**ABSTRACT**

Artificial intelligence (AI) has huge potential to improve the health and well-being of people, but adoption in clinical practice is still limited. Lack of transparency is identified as one of the main barriers to implementation, as clinicians should be confident the AI system can be trusted. Explainable AI has the potential to overcome this issue and can be a step towards trustworthy AI. In this paper we review the recent literature to provide guidance to researchers and practitioners on the design of explainable AI systems for the health-care domain and contribute to formalization of the field of explainable AI. We argue the reason to demand explainability determines what should be explained as this determines the relative importance of the properties of explainability (i.e. interpretability and fidelity). Based on this, we propose a framework to guide the choice between classes of explainable AI methods (explainable modelling versus post-hoc explanation; model-based, attribution-based, or example-based explanations; global and local explanations). Furthermore, we find that quantitative evaluation metrics, which are important for objective standardized evaluation, are still lacking for some properties (e.g. clarity) and types of explanations (e.g. example-based methods). We conclude that explainable modelling can contribute to trustworthy AI, but the benefits of explainability still need to be proven in practice and complementary measures might be needed to create trustworthy AI in health care (e.g. reporting data quality, performing extensive (external) validation, and regulation).

**Highlights**
- Comprehensive survey to provide guidance and formalize the field of explainable AI
- Assessment of quantitative evaluation metrics for explainability
- Step-by-step guidance to choose between classes of explainable AI methods
- Explainable AI can contribute to trustworthy AI, but other measures might be needed

## 1. Introduction

Artificial intelligence (AI) offers great opportunities for progress and innovation due to its ability to solve cognitive problems normally requiring human intelligence. Practical successes of AI in a variety of domains already influenced people's lives (e.g. voice recognition, recommendation systems, and self-driving cars). In the future, AI is likely to play an even more prominent role. The International Data Corporation estimates spending on AI to increase from 37.5 billion in 2019 to 97.9 billion in 2023 [1]. Due to the increasing availability of electronic health records (EHRs) and other patient-related data, AI also has huge potential to improve the health and well-being of people. For example, by augmenting the work of clinicians in the diagnostic process, signaling opportunities for prevention, and providing personalized treatment recommendations. Although some simple assistive tools have been deployed in practice [2, 3], there is no widespread use of AI in health care yet [4, 5].

Lack of transparency is identified as one of the key barriers to implementation [5, 6]. As it is the responsibility of clinicians to give the best care to each patient, they should be confident that AI systems (i.e. AI models and all other parts of the implementation) can be trusted. Health care is a domain with unique ethical, legal, and regulatory challenges as decisions can have immediate impact on the well-being or life of people [7]. Often-mentioned concerns include potential algorithmic bias and lack of model robustness or generalizability. Other problems include the inability to explain the decision-making progress of the AI system to physicians and patients, difficulty to assign accountability for mistakes, and vulnerability to malicious attacks. It is still unclear how to implement and regulate trustworthy AI systems in practice. We follow the definition of the High-Level Expert Group on AI[1] that *trustworthy AI* should satisfy three necessary conditions: AI systems should comply with all applicable laws and regulations (lawfulness), adhere to ethical principles and values (ethicality), and be safe, secure and reliable (robustness) [8]. It is difficult to ensure that these conditions hold as there are no proven methods to translate these conditions into practice [9].

---
[1] The High-Level Expert Group on AI is an independent group of experts set up by the European Commission.

A possible step towards trustworthy AI is to develop *explainable AI*. The field of explainable AI aims to create insight into how and why AI models produce predictions, while maintaining high predictive performance levels. In a recent report, the European Institute of Innovation and Technology Health [10] identified 'explainable, causal and ethical AI' as a potential key driver of adoption. Other guidelines mention explainability as a requirement of trustworthy AI [8, 11]. Although the field of explainable AI has promising prospects for health care, it is not fully developed yet. Among others, it is unclear what a suitable explanation is and how its quality should be evaluated. Moreover, the value of explainable AI methods remains to be proven in practice.

Previous research on explainable AI includes work on formal definitions [12, 13*], development of explainable AI techniques (for an extensive overview see Guidotti et al. [14]), and - to a lesser extent - evaluation methods (for a recent survey see Mohseni, Zarei and Ragan [15*]). For high-level introductions to the field we refer to [16-18]. Murdoch, Singh, Kumbier, Abbasi-Asl and Yu [19] present a common vocabulary to help select and evaluate explainable AI methods. For a domain-specific introduction, we point to Ahmad et al. [7] who reviewed the notion of explainability and its challenges in the context of health care. For a recent review focusing on applications of explainable AI models in health care we refer to Payrovnaziri et al. [20]. Open problems include: a remaining lack of agreement on what explainability means [3, 14, 16], no clear guidance how to choose amongst explainable AI methods [19], and the absence of standardized evaluation methods [14, 16, 19].

In this paper we investigate how explainable AI can contribute to the bigger goal of creating trustworthy AI. We chose not to perform a systematic review of the literature, therefore not following the PRISMA guidelines [21], since we felt that the heterogeneous and often ill-defined terminology in the field of explainable AI prevented a focused search strategy. Rather, we reviewed the last five years of literature in the field of explainable AI, searching for papers that present conceptual frameworks or methodology, not applications of existing techniques. Our search terms included 'explainability' or 'interpretability' combined with 'artificial intelligence', 'machine learning' or 'black box'. Papers were collected from various sources such as PubMed, ScienceDirect, IEEE Xplore, ACM Digital Library, and Google Scholar. We identified influential papers presenting definitions of explainability and related terms, reviewing or structuring the field of explainable AI, or presenting (other) novel ideas. We further completed this set using the reference lists of the selected papers and by investigating citations to these papers. As we are interested in the recent advancement in the field, we also included preprints posted on arXiv. Preprints are marked by * to distinguish them from peer-reviewed literature. We wanted to answer the following questions:

- What does explainability mean? (Section 2)
- Why and when can explainability be useful? (Section 3)
- Which explainable AI methods are available? (Section 4)
- How can explainability be evaluated? (Section 5)
- How to choose amongst different explainable AI methods? (Section 6)

By answering these questions we aim to provide guidance to researchers and practitioners on the design of explainable AI systems for the health-care domain and contribute to formalization of the field of explainable AI.

## 2. What does explainability mean?

In this section we discuss the meaning of the terms explainability, interpretability, comprehensibility, intelligibility, transparency, and understandability in more detail. Lipton [12] points out that these terms are often ill-defined in the existing literature. Researchers do not specify what terms mean, use the same term for different meanings, or refer to the same notion by different terms. Terms are used differently in the public versus scientific setting [16] and across AI communities [22*]. There is a widely recognized need for a more formal definition of the properties that explanations should satisfy [3, 14, 16].

Figure 1 presents a summary of the proposed definitions. In the remainder of this section we elaborate on these definitions and discuss alternative definitions used in the literature.

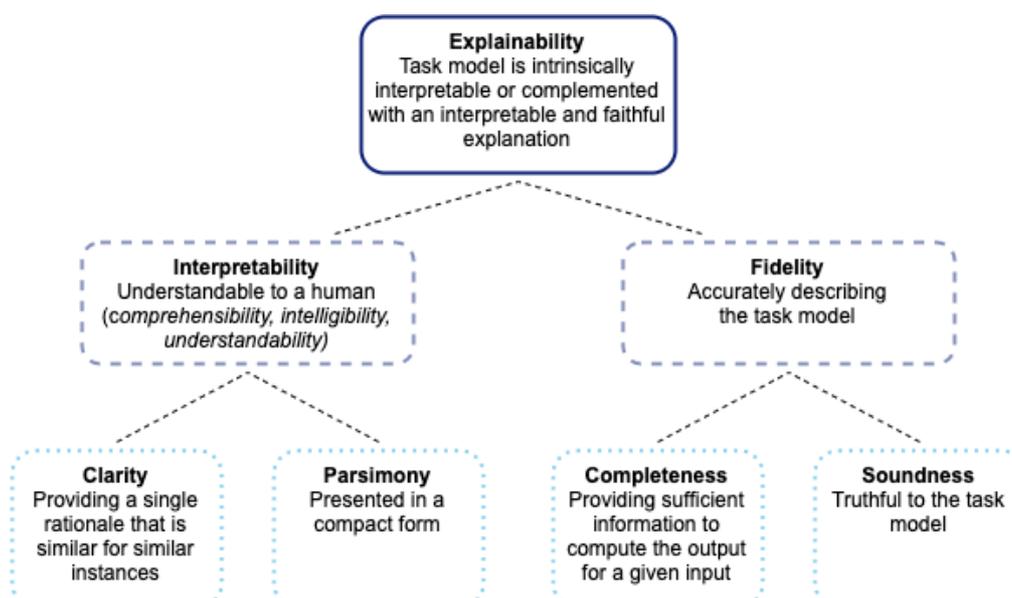

Figure 1: Proposed definitions for explainability and related terms. Explainability is the overarching concept, consisting of different properties.

## 2.1 Explainability

Some use explainability and interpretability synonymously [23, 24]. However, looking at the literature, we suggest to follow Gilpin et al. [25], who state that *interpretability* and *fidelity*[2] are both necessary to reach *explainability*. The interpretability of an explanation captures how understandable an explanation is for humans. The fidelity of an explanation expresses how accurately an explanation describes model behavior, i.e. how faithful an explanation is to the task model. Hence, they argue an explanation should be understandable to humans and correctly describe model behavior in the entire feature space. We propose this definition of explainability because we believe fidelity is a crucial term that should be taken into account in the design of explainable AI systems. The importance of a faithful explanation is also recognized by others (e.g. [26]). It is a challenge in explainable AI to achieve both interpretability and fidelity simultaneously. In line with Arrieta et al. [17] we consider interpretability a property related to an explanation and explainability a broader concept referring to all actions to explain. The explanation can be the task model or a post-hoc explanation. The *task model* is the model generating predictions. A *post-hoc explanation* accompanies an AI model and provides insights without knowing the mechanisms by which the model works (e.g. by showing feature importance). This leads to the following definition:

> Definition 1: explainability
>
> An AI system is explainable if the task model is intrinsically interpretable (here the AI system is the task model)[3] or if the non-interpretable task model is complemented with an interpretable and faithful explanation (here the AI system also contains a post-hoc explanation).

We discuss different explainable AI methods in Section 4.

## 2.2 Interpretability and fidelity

For interpretability, various definitions exist in the literature. We distinguish three types of definitions:

1. Definitions based on formal aspects of system operations. For example, in the computer science field, a system is considered interpretable if the relation between input and output can be formally proven to be correct [27*]. A common objection to this type of definitions is that they do not focus enough on the user value of explanations [7].

---

[2] Note that the term used by Gilpin et al. [25] is completeness, but we use the term fidelity throughout this paper as it is more commonly used in the literature.
[3] In the case of an intrinsically interpretable model, fidelity is guaranteed by design.

2. Definitions focused on the explanatory value to the user. One commonly used definition is provided by Doshi-Velez and Kim [13*], who define interpretability as the ability to explain or to present AI systems in understandable terms to a human. Although these definitions give the user a central role, a problem with this type of definitions is that it is still not clearly defined what it means to be understandable.

3. Definitions viewing interpretability as a latent property. Poursabzi-Sangdeh, Goldstein, Hofman, Vaughan and Wallach [28*] argue that interpretability cannot directly be observed and measured. Instead, they define interpretability as the collection of underlying manipulatable factors influencing model complexity (e.g. number of features, model representation) that play a role in influencing different outcomes of interest.

To achieve a set of practical definitions, we split both interpretability and fidelity into generic underlying factors that together determine the quality of an explanation (definition type 3). This leads to the following definitions:

Definition 2: interpretability
An explanation is interpretable if [29]:
   a. the explanation is unambiguous, i.e. it provides a single rationale that is similar for similar instances (*clarity*),
   b. the explanation is not too complex, i.e. it is presented in a compact form (*parsimony*).
Interpretability describes the extent to which a human can understand an explanation.

Definition 3: fidelity
An explanation is faithful if [30]:
   a. the explanation describes the entire dynamic of the task model, i.e. it provides sufficient information to compute the output for a given input (*completeness*)[4],
   b. the explanation is correct, i.e. it is truthful to the task model (*soundness*).
Fidelity describes the descriptive accuracy of an explanation.

---

[4] The definition of completeness by Kulesza et al. [30] used in this paper is more specific than that of Gilpin et al. [25].

In Section 5 we investigate how to evaluate each of these properties quantitatively for different types of explainable AI methods. It is important to note that the usefulness of an explanation is influenced by the expertise (the level of AI or domain knowledge), preferences, and other contextual values of the target user [16]. The terms defined above thus depend on the user, which could be a developer, deployer (owner), or end-user (e.g. a clinician) of an AI application. Moreover, the usefulness of an explanation depends on the reason to demand explainability (more on this in Sections 3 and 6) [25, 27*].

### 2.3    Related terms

Other common terms in the literature are comprehensibility, intelligibility, transparency, and understandability. As with explainability and interpretability, some equate these to other terms, whereas others attach different meanings. Guidotti et al. [14] use comprehensibility as a synonym for interpretability. Doran et al. [22*] distinguish interpretability and comprehensibility by the presence (or absence) of transparency. They define a model to be transparent if the inner workings of the model are visible and one can understand how inputs are mathematically mapped to outputs. The opposite of a transparent model is an opaque model. They argue an interpretable model is transparent, but a comprehensible model can be opaque. We do not consider this distinction necessary from a practical viewpoint. The inaccessibility of a model's inner workings limits how accurately an explanation can describe the task model and is thus captured in fidelity. Hence, like Arrieta et al. [17] we say a model is *transparent* if it is by itself interpretable. Intelligibility is sometimes referred to as the degree to which a human can predict how a change in the AI system will affect the output [31*]. Others equate intelligibility, understandability and interpretability [23, 32]. We view intelligibility as one possible way to measure interpretability, but do not use this as separate notion as it does not describe a different goal. We thus do not distinguish comprehensibility, intelligibility, and understandability from interpretability.

### 3.  Why and when can explainability be useful?

Some argue explainable AI is necessary for the field of AI to further develop [16]. However, the importance of explainability depends on the application domain and specific use case. In this section we explore why and when explainability can be useful in health care. Diverse reasons to demand explainability are mentioned in the literature [12, 13*, 31*]. Some mention potential problems of AI models that could be detected by explanations (e.g. use of a wrong or incomplete objective, dataset shift), others refer to model desiderata (e.g. reliability, legality) or end-goals (e.g. enhancing user acceptance, building trust).

Adadi and Berrada [16] summarize the literature by formulating four motivations for explainability: 1) to justify decisions and comply with the 'right to explanation', 2) to enable user control by identifying and correcting mistakes, 3) to help improve models by knowing why a certain output was produced, and 4) to gain new insights by investigating learned prediction strategies. However, this taxonomy does not mention one commonly mentioned motivation for explainability, as a means to verify whether other model desiderata are satisfied [13*]. Examples of such model desiderata are fairness (i.e. protected groups are not discriminated against), generalizability (i.e. model transferable to different data), privacy (i.e. sensitive information in data protected), robustness (i.e. performance independent of input data), and security (i.e. model not susceptible to adversarial attacks). Although model desiderata are often used to motivate explainability, we stress explainability is not necessary to fulfill these model desiderata. Moreover, the benefits of explainability still need to be proven in practice.

We classify explainable AI systems based on the need they strive to fulfill, as this determines the relative importance of the properties of explainability and thus influences the design choice of explainable AI systems. We distinguish three reasons why explainability can be useful:

1. To assist in verifying (or improving) other model desiderata. Clinical tasks are often complex and it is impossible to incorporate and evaluate all desirable properties (e.g. legality, ethicality, and robustness) in a model using standard predictive performance evaluation metrics (e.g. area under the receiver-operating curve) [12]. Explanations can help by allowing the possibility of a human in the loop to detect and correct problems in some situations. For example, researchers interviewing clinicians found that explanations describing which features were used to predict the outcome, are desired by clinicians to verify whether (in)appropriate features were used by the model [3]. Although using explainability to verify other model desiderata is mentioned more commonly in the literature, we note that explanations cannot guarantee that model desiderata are satisfied.

2. To manage social interaction. The need for explainability can also be motivated by the social dimension of explanations [33]. One reason why people generally ask for explanations is to create a shared meaning of the decision-making process. This is important in the health-care domain to comply with the 'right to explanation' in the EU General Data Protection Regulation (GDPR) of the European Union [34]. Even when there is no legal obligation, it is important for clinicians to be able to justify their decision making towards colleagues and patients [3].

3. To discover new insights. One can also demand explainability to learn from the models for knowledge discovery. Explainability enables comparisons of learned strategies with existing knowledge and facilitates learning for educational and research purposes. These insights can be used to guide future research, e.g. to help with new drug development or to design clinical trials.

However, explanations can be costly (time consuming to design and to use) and might only be needed in some situations. First, explanations might be needed when the cost of misclassification is high, for example, in safety-critical applications where life and health of humans is involved or when there is potential of huge monetary losses. Second, explanations can be required when the AI system has not yet proven to work well in practice and we still need to work on user trust, satisfaction, and acceptance. In the health-care domain both situations apply; stakes can be high and it may be difficult to test AI systems to a satisfactory level before deploying them in real-life. However, when the model has no significant impact or has proven its performance sufficiently, the cost of explanation may outweigh the benefit [35].

### 4. Which explainable AI methods are available?

There are many different explainable AI methods described in the literature. One way to achieve explainable AI is by *explainable modelling*, i.e. by developing an AI model where the internal functioning is directly accessible to the user, so that the model is intrinsically interpretable. Alternatively, *post-hoc explanations* can accompany the AI model to make it explainable. Post-hoc explanations can be motivated by the sometimes occurring trade-off between predictive performance and interpretability. Hence, instead of developing an intrinsically interpretable model with the risk of a lower predictive performance, post-hoc explanations accompany the AI model and provide insights without knowing how the AI model works. Post-hoc explanation methods can be classified as model-agnostic (can be used to explain any kind of model) or model-specific (only suitable for specific model classes). In addition, some provide global explanations and others local explanations. The difference between these two lies in the scope of explanation, either providing a rationale for an individual prediction or for the model as a whole. Global explanations could also be used to explain individual predictions, but are less accurate than local explanations.

Different taxonomies have been proposed based on the explanation-generating mechanism, the type of explanation, the scope of explanation, the type of model it can explain, or a combination of these features [14, 36*]. We classify explainable AI techniques according to the type of explanation and the scope of explanation, as we believe this is the relevant distinction when choosing the most appropriate explainable AI method. We distinguish three types of explanations: *model-based explanations*, *attribution-based explanations*, and *example-based explanations*. Each type of explanation can be used to provide a global or local explanation. The resulting classes differ in the information they provide and thus score differently on the properties of explainability. In explainable modelling there is no difference in scope, as the task model gives both explanations. We focus on post-hoc explanation methods that are model-agnostic. The explainable AI methods discussed in this section are summarized in Table 1. We now discuss each type of explanation in more detail.

Table 1: Proposed classification of explainable AI methods with examples.

| Approach | Type of explanation | Scope | Examples of explainable AI methods |
|---|---|---|---|
| Explainable modelling | Model | | Adopt intrinsically interpretable model, architectural modifications (regularization), developing hybrid models, or training the task model to provide explanations |
| Post-hoc explanation | Model | Global | BETA - Lakkaraju, Kamar, Caruana and Leskovec [37*]<br>Tree extraction - Bastani, Kim and Bastani [38*]<br>Distill-and-compare - Tan, Caruana, Hooker and Lou [39]<br>Symbolic metamodeling - Alaa and van der Schaar [40] |
| | | Local | LIME - Ribeiro et al. [26]<br>Anchors - Ribeiro, Singh and Guestrin [41] |
| | Attribution | Global | PDP - Friedman [42]<br>Feature interaction - Friedman and Popescu [43]<br>ALE - Apley and Zhu [44*]<br>Feature importance - Fisher, Rudin and Dominici [45*]<br>LOCO - Lei, G'Sell, Rinaldo, Tibshirani and Wasserman [46] |
| | | Local | ICE - Goldstein, Kapelner, Bleich and Pitkin [47]<br>QII - Datta, Sen and Zick [48]<br>SHAP - Lundberg and Lee [49]<br>LOCO - Lei et al. [46]<br>INVASE - Yoon, Jordon and van der Schaar [50] |
| | Example | Global | Influential instances - Cook [51]<br>MMD-critic - Kim, Khanna and Koyejo [52] |
| | | Local | influential instances - Cook [51]<br>Unconditional counterfactual explanations - Wachter, Mittelstadt and Russell [53] |

### 4.1 Model-based explanations

This class includes all methods that use a model to explain the task model. Model-based explanations fall under explainable modelling as well as post-hoc explanations. Either the task model itself is used as explanation (explainable modelling) or another, more interpretable model is created to explain the task model (post-hoc explanation). Explainable modelling provides full transparency of the decision-making process of the model and is preferred if the model is simple enough to be interpreted by clinicians. A post-hoc model-based explanation can also be valuable in health care, but has the disadvantage that it might not be faithful and clinicians need to understand this limitation. Note that the task model can always serve as a first explanation to the user, even when other post-hoc explanations are used.

In explainable modelling, the aim is to develop a task model that is by itself interpretable for the user. To develop an intrinsically interpretable model, one can opt for a model class that is known to generate interpretable models for humans. Three model classes that are typically considered interpretable are sparse linear classifiers (e.g. linear/logistic regression, generalized additive models), discretization methods (e.g. rule-based learners, decision trees), and example-based models (e.g. K-nearest neighbors) [54]. However, interpretability is also influenced by other factors such as the number and comprehensibility of input features. Hence, even though a decision tree is typically considered easier to interpret than a neural network, a deep decision tree may be less interpretable than a compact neural network. Rule-based learners and logistic regression models are both commonly used in AI systems designed for health care. As an alternative, Caruana et al. [55] investigated the use of generalized additive models (GAMs) with pairwise interactions to predict pneumonia risk and 30-day hospital readmission. This interpretable model allowed them to recognize and correct the erroneous outcome that having asthma lowers the risk of dying from pneumonia. This could also be done with rule-based learners and logistic regression models, but the GAM with pairwise interactions gave the best overall predictive performance. Other ways to obtain an interpretable model include using architectural modifications (e.g. some form of regularization) [56, 57, 58*], developing hybrid models that combine interpretable and black box (i.e. non-interpretable) models [59], or training the task model with an annotated dataset to provide explanations [60].

Alternatively, we can distill a more interpretable surrogate model that approximates the task model and use this model in the form of a post-hoc explanation. This can give new insights by revealing important features and interactions between them. Lakkaraju et al. [37*] present a methodology called black box explanations through transparent approximations (BETA) to derive a global surrogate model. Their method learns a compact set of decision rules, by jointly optimizing unambiguity, fidelity, and interpretability. Che, Purushotham, Khemani and Liu [61] used mimic learning to transfer knowledge from deep neural networks to smaller, more interpretable models for intensive care unit outcome prediction to verify agreement with earlier findings. Furthermore, Bastani et al. [38*] introduced a novel algorithm to learn a representation in the form of a decision tree and use this to predict the risk of type II diabetes. The interpretability of the decision tree enabled them to analyze differences in results between patients from different providers and the effect of certain features (e.g. age). More recently, Alaa and van der Schaar [40] predicted the risk of mortality for breast cancer patients applying their newly proposed symbolic metamodeling framework using Meijer G-function parametrization, minimizing the metamodeling loss via gradient descent. This metamodel can give insight in the impact of interactions (e.g. between estrogen-receptor status, number of nodes, and tumor size) on a patient's risk. Instead of serving as explanation, global surrogate models can also replace the task model [7]. LIME (local interpretable model-agnostic explanations) and Anchors are examples of local surrogate models. Sometimes these methods are classified as attribution methods (e.g. by Guidotti et al. [14]), because the resulting local surrogate models are used to derive the important input features. However, model-based explanations can always be used to generate other types of explanations and as both methods output local models we argue they belong here.

### 4.2 Attribution-based explanations

Attribution methods rank or measure the explanatory power of input features and use this to explain the task model. These methods are sometimes also called feature/variable importance, relevance, or influence methods. The majority of post-hoc explanation methods fall in this class. For clinicians, this type of explanation is helpful to learn which features are responsible for the predicted outcome and to be able to compare with their own prior knowledge in limited time [3]. We can classify attribution methods according to the explanation-generating mechanism into perturbation and backpropagation methods [62*]. Methods based on backpropagation are not model-agnostic, as they are often designed for a specific model class or require the model function to be differentiable. We thus only discuss methods based on perturbation here.

Two examples of visual global attribution methods are Partial Dependency Plots (PDPs) [45*] and Accumulated Local Effects (ALE) plots [44*], both show the average effect of input features to the output. Alternatively, Feature Importance [45*] measures the increase in prediction error after permuting a feature, and Feature Interaction (or H-statistic) [43] calculates the amount of variation explained by interaction terms. Other methods exists to generate local explanations. Individual Conditional Expectation (ICE) plots show the expected model prediction as a function of feature values for a given instance [47]. Inspired from game theory, Shapley values indicate how to fairly distribute importance ('payout') across features ('players') by averaging the marginal contribution across all possible coalitions. Quantitative Input Influence (QII) is one sampling approximation to compute Shapley values [48]. The SHAP method is another way to compute these values [49]. Next, INVASE can determine a flexible number of relevant features for each instance using a model consisting of a selector, predictor, and baseline network [50]. Finally, Lei et al. [46] introduced a global and local measure of variable importance based on Leave-One-Covariate-Out (LOCO) inference.

Attribution-based methods are commonly used in health care. For example, LIME has been used to determine if identified important features are in line with prior knowledge. In the study by Pan et al. [63] this helped to show that the most important factors in the clinic to diagnose potential central precocious puberty (i.e. basal serum luteinizing hormone, insulin-like growth factor-I, and follicle-stimulating hormone) were incorporated in the model. Similarly, the important features of a model to predict intensive care mortality were found to be consistent with medical understanding [64*]. However, LIME has also been used to explore determinants of an outcome. For example, Ghafouri-Fard et al. [65] used it to identify the most protective and risky genotypes for autism spectrum disorders.

### 4.3    Example-based explanations

The methods in this class explain the task model by selecting instances from the dataset or creating new instances, e.g. by selecting prototypes (i.e. instances that are well predicted by the model) and criticisms (i.e. instances that are not well predicted by the model), identifying influential instances for the model parameters or output, or creating a counterfactual explanation. These methods find or create similar patients to explain and can help clinicians if instances can be represented in a human-understandable way. There are relatively few methods available in the literature of this class.

Kim et al. [52] developed MMD-critic to learn prototypes and criticisms for a given dataset using the maximum mean discrepancy (MMD) as a measure of similarity. A method to compute influential instances was developed by Cook [51], who uses deletion diagnostics to measure the effect of deleting an instance on the model predictions. Influential instances can provide both global and local level explanations. Finally, counterfactual explanations require hypothetical thinking of the form: 'If X had not occurred, Y would not have occurred'. These can be generated by trial and error, but Wachter et al. [53] present a more sophisticated approach to generate unconditional counterfactual explanations. They demonstrate their explanations on a case study predicting the risk of diabetes, which results in explanations such as 'if your 2-hour serum insulin level was 154.3, you would have a score of 0.51' (i.e., have a 51 percent risk of diabetes) [53]. This type of explanation is easy to communicate between clinicians and patients and similar to how clinicians are used to communicate (e.g. 'if your body mass is above 30, you are obese').

### 5.    How can explainability be evaluated?

Many papers claim to have reached their goal without formal evaluation ('I know it when I see it') [13*]. Several researchers stress the need for formal evaluation metrics and a more systematic evaluation of the methods [14, 16]. Before discussing the relevant literature, we point out that the goal of evaluation methods is twofold. First, evaluation allows a formal comparison of available explanation methods. Many methods have been proposed, often with a similar goal, but it is unclear which one is to be preferred. When evaluating post-hoc explanations, the problem is there is no ground truth, as we do not know the real inner workings of the model [33]. Second, evaluation offers a formal method to assess if explainability is achieved in an application. Here the focus lies on determining if the offered form of explainability achieves the defined objective [12].

Doshi-Velez and Kim [13*] divide evaluation approaches in *application-grounded* (experiments with end-users), *human-grounded* (experiments with lay humans), and *functionality-grounded evaluation* (proxies based on a formal definition of interpretability). Application- and human-grounded evaluations are less objective than functionality-grounded evaluation as the results of the former methods depend on the selected pool of humans. Within human experiments, one can use both qualitative and quantitative metrics. Qualitative metrics include asking about the usefulness of, satisfaction with, and trust in provided explanations by means of interviews or questionnaires [66, 67*]. Quantitative metrics include measuring human-machine task performance in terms of accuracy, response time needed, likelihood to deviate, or ability to detect errors [28*, 66, 67*, 68*].

Although application-grounded evaluation approaches provide the strongest evidence of success [13*], developing an AI system by repeatedly updating and evaluating on humans may be an inefficient process. Experiments with humans are necessary and provide valuable information, but are expensive, time-consuming, and subjective. For a comprehensive survey of (qualitative) evaluation measures to assess different design goals (e.g. user mental model, user trust, and human-machine task performance), we refer to the recent survey by Mohseni et al. [15*]. In the remainder, we focus on quantitative metrics, which are less well-studied in the literature and are needed for an objective initial assessment of explanation quality and a formal comparison of explanation methods [33].

In the remainder of this section we evaluate to what extent the properties of explainability, i.e. interpretability (consisting of clarity and parsimony) and fidelity (consisting of completeness and soundness), are satisfied for model-based, attribution-based, and example-based explanations. In particular, we discuss which properties are satisfied by definition, and which quantitative proxy metrics are available to evaluate the remaining properties. We focus on model-agnostic evaluation methods that are domain and task independent.

### 5.1    *Evaluating model-based explanations*

Starting with fidelity, we know that model-based explanations provide sufficient information to compute the output for a given input and thus always satisfy the completeness property. Similarly, we know that soundness is satisfied when the task model itself is used as explanation, as is the case in explainable modelling. For post-hoc model-based explanations, we can measure soundness using the fidelity metric of Lakkaraju et al. [37*]. They define fidelity as the level of (dis)agreement between the task model and post-hoc model explanation and measure the percentage of predictions that are the same.

Continuing with interpretability, we argue that global explanations usually satisfy the clarity property, as these methods provide one rationale for the entire model. Hence, global model-based, attribution-based, and example-based methods satisfy clarity. An exception to this are model-based explanations that can provide multiple rationales because of ambiguity in the model itself, for example, due to overlapping rules in the case of an unordered rule-based system. Lakkaraju et al. [37*] defined a measure of unambiguity based on rule overlap and coverage of feature space that can be used to quantify clarity in this specific case. Local explanations, on the other hand, have the problem that explanations can be very different for similar instances. We did not find a metric to assess the clarity of local model-based explanations.

Finally, we found that most evaluation methods for model-based explanations focus on parsimony. Model size is often used as an approximation for model complexity and used to measure the level of model interpretability [14, 69*]. Examples of metrics are the number of features (e.g. non-zero weights, number of rules, or features used in splits) and the complexity of relations (e.g. interactions, length of rules, tree depth), but these metrics are model dependent. Recently, Friedler et al. [68*] proposed to measure model complexity by the number of runtime operation counts (i.e. the number of boolean and arithmetic operations) needed to run the model for a given input, and investigated the relation with the accuracy and time needed to perform tasks. Molnar et al. [69*] used a different approach and quantified model complexity using the number of features used, interaction strength, and main effect complexity. Hence, we conclude that (multiple) metrics to measure parsimony are available for model-based explanations.

### 5.2    *Evaluating attribution-based explanations*

Again starting with fidelity, we conclude that attribution-based explanations only provide a partial explanation of the model and thus do not satisfy completeness. We did not find metrics to assess the completeness of attribution-based explanations. However, different methods to evaluate attribution-based explanations have been proposed in the literature and we conclude that the majority can be used to measure soundness.

We distinguish between *empirical* and *axiomatic* evaluation approaches. Empirical evaluation approaches directly measure the performance of the attribution method. Some do this by artificially creating a ground truth for evaluation [26, 70*]. Other methods are based on measuring performance degradation using perturbation analysis [71, 72]. Axiomatic evaluation approaches define what an ideal attribution method should achieve and evaluate whether this is accomplished. Several researchers adopt axiomatic approaches, evaluating whether certain basic properties of explanations hold. Kindermans and Hooker propose a test to investigate if an explanation is sensitive to a meaningless *transformation of the input data* [33]. Sundararajan, Taly and Yan [73] investigate two other axioms: *sensitivity* and *implementation invariance*. They define an attribution method as sensitive when a feature has a non-zero attribution if changing that feature for a given baseline results in different predictions. Similarly, an attribution method is implementation invariant if it always produces identical attributions for functionally equivalent models (i.e. models producing the same output for the same input). Montavon, Samek and Müller [74] discuss explanation *continuity* and *selectivity* as important explanation properties. Explanation continuity means that if two nearly identical data points have a nearly identical model response, the corresponding explanations should also be nearly identical. Furthermore, they define selectivity as the ability of an explanation to give relevance to variables that have the strongest impact on the prediction value. Finally, Ancona et al. [62*] state an attribution method satisfies *sensitivity-N* when the sum of the attributions for any subset of features of cardinality N is equal to the variation of the output caused by removing the features in the subset. This property is also called *completeness* or *summation to delta* [33, 73]. Tests for these properties can be used to assess the soundness of explanations, with the exception of continuity [74], which relates to the clarity of an explanation. Hence, we conclude that metrics for soundness of attribution-based explanations and clarity of local attribution-based explanations are available.

Unfortunately, both methods have important limitations. Although empirical evaluation approaches capture the essence of what an attribution method should achieve, they cannot distinguish the performance of the task model from the performance of the attribution method [73]. Axiomatic evaluation approaches, on the other hand, have the disadvantage that it is difficult to define what we expect from attribution methods. More research is necessary to define the desired behavior of explanation methods and to translate them to testable properties [33]. Finally, we argue that parsimony is satisfied for attribution-based methods as long as the feature itself is understandable, because the information load of these methods is limited.

### 5.3    Evaluating example-based explanations

We do not know of any example-based evaluation methods. Evaluation methods for example-based explanations might be underdeveloped, because the methods have slightly different objectives (i.e. selecting prototypes and criticisms, identifying influential instances, or creating counterfactual explanations) and less methods are available of this class. Hence, using similar reasoning as before we conclude that clarity is satisfied for global example-based explanations and example-based explanations are parsimonious as long as the instance itself is understandable. The remaining properties are not satisfied and no evaluation methods were found.

The findings in this section are summarized in Table 2.

Table 2: Evaluation of properties of explainability for different explainable AI methods.

| Approach | Type of explanation | Scope | Interpretability | | Fidelity | |
|---|---|---|---|---|---|---|
| | | | Clarity | Parsimony | Completeness | Soundness |
| Explainable modelling | Model | | S* | M | S | S |
| Post-hoc explanation | Model | Global | S* | M | S | M |
| | | Local | U | M | S | M |
| | Attribution | Global | S | S** | U | M |
| | | Local | M | S** | U | M |
| | Example | Global | S | S** | U | U |
| | | Local | U | S** | U | U |

For each explainable AI method we indicate if a property is by definition satisfied (S), if a property can be evaluated using a generic quantitative metric (M), or if metrics are unavailable (U). *Satisfied if model cannot provide multiple rationales. **Satisfied if feature or instance is understandable for a human.

## 6.    How to choose amongst different explainable AI methods?

How to best design explainable AI systems is a non-trivial problem. This section provides guidance for developers that design AI systems for health care or other domains where explainability can be useful. We argue that different properties of explainability can be more or less important (see Section 2) depending on the reason to demand explainability (see Section 3). In this section, we discuss the trade-offs that occur while designing AI systems. Based on the classification of explainable AI methods and the findings in the previous section (summarized in Table 2), we propose the step-by-step guide in Figure 2 to select the most appropriate class of explainable AI methods. This can be used by developers to systematically think through and discuss the relevant trade-offs with the target user of the AI system. We will now discuss each of the steps.

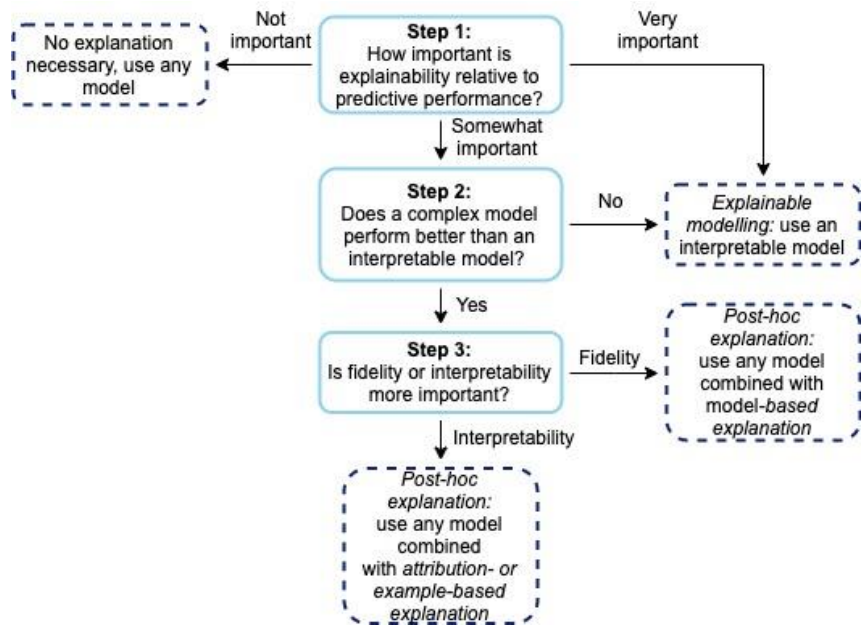

Figure 2: Proposed framework with concrete design recommendations to choose amongst explainable AI methods.

### 6.1 *Step 1: How important is explainability relative to predictive performance?*

Predictive performance is of crucial importance. No one would be willing to adopt an AI system that has unsatisfactory performance. On the contrary, the importance of explainability depends on the application domain and specific use case (see Section 3). If explainability is not important and it is acceptable to have a black box model, it is best to look for the model with the best predictive performance as explanations can be costly.

When explainability is (somewhat or very) important, one needs to choose amongst explainable AI methods. Table 2 showed it is impossible to satisfy (or even quantify) all properties of explainability for all explainable AI methods. If one cares about explainability, one should choose between two approaches: explainable modelling or post-hoc explanation. Choosing explainable modelling may mean giving up on predictive performance, choosing post-hoc explanation means giving up some explainability. As the developer of an AI system, it is thus important to establish the relative importance of explainability compared to predictive performance and what is desired by end-users of the AI system.

Developers should be aware that post-hoc explanations are approximations of the task model's inner workings and are by definition not completely faithful. These explanations have the potential to present plausible but misleading explanations [12, 75]. As the goals of the explainer and the user of the explanations are not necessarily aligned, it is difficult for a user to determine whether this type of explanation can be trusted. Whereas the goal of the explainer could be to simply generate user trust (i.e. might be beneficial not to reveal mistakes), the user might want to understand limitations of the AI system (i.e. is interested in mistakes). Moreover, explainers can provide an empty explanation (i.e. without information content) to soothe users or carefully select one suiting their goals [33]. Hence, in cases where both interpretability and fidelity are very important, i.e. explainability is required, explainable modelling is the most appropriate design choice. When it is established that explainability is somewhat important, the next relevant question for developers is whether a complex model is performing better than an interpretable model.

### 6.2   *Step 2: Does a complex model perform better than an interpretable model?*

Machine learning algorithms that are known for their promising predictive performance are often not interpretable (e.g. neural networks) and vice versa (e.g. linear regression). Hence, a trade-off between predictive performance and interpretability might exist. However, it is important to note that this trade-off does not always occur, in which case a more interpretable model is generally preferred (i.e. explainable modelling) [76]. As the developer of an AI system, it is thus important to not only investigate the use of complex models, but to also compare this to an interpretable alternative. If the predictive performance does decrease substantially when employing an interpretable model, one can opt for a post-hoc explanation.

### 6.3   *Step 3: Is fidelity or interpretability more important?*

The benefit of choosing for post-hoc explanation, is that the model with the best predictive performance can be used as task model (regardless of its interpretability). However, when choosing a post-hoc explanation a new trade-off arises, between interpretability and fidelity. As the developer of an AI system, it is thus important to determine which property is most important for the current application. In general, we believe fidelity is most important to assist in verifying other model desiderata or to discover new insights, as it is essential to find the true underlying mechanisms of the model. In this case, we argue that model-based explanations are most suitable, as we concluded in Section 5 that model-based explanations satisfy completeness. We argue that interpretability, on the other hand, is most important to manage social interaction, as we know from social sciences that humans also tailor their explanation to their audience and do not necessarily give the most likely explanation [24]. In this case, attribution- and example-based explanations, are often considered better as these are more parsimonious. The choice between local and global explanations is another consideration that depends on the use case and needs to be made by the developer of an AI system. Whereas global explanations are attractive as they usually satisfy clarity, local explanations might often be more appropriate in a health-care setting.

## 7. Discussion

Strategies to develop and regulate trustworthy AI systems are still under development [8, 11, 77]. As the demands for explainable AI and trustworthy AI are closely related, we investigated the role of explainability in creating trustworthy AI. We extended other recent surveys [14, 15*, 16-19] by providing a framework with concrete recommendations to choose between classes of explainable AI methods. Furthermore, we proposed practical definitions and contributed to the existing literature by assessing the current state of quantitative evaluation metrics. We now discuss how explainable AI can contribute to the bigger goal of creating trustworthy AI, and then highlight other measures to create trustworthy AI in health care.

### 7.1    Developing explainable AI to create trustworthy AI

In Section 6, we argued that the reason to demand explainability determines what should be explained and we proposed a step-by-step guide with concrete design recommendations for developers. We hope this framework provides the groundwork necessary to design explainable AI systems that are useful in practice. When applied to make design choices for trustworthy AI, we note that ensuring the AI system is lawful, ethical, and robust coincides with the reason to assist in verifying other model desiderata. For this need, we believe explainability - and especially fidelity - is extremely important. Hence, our framework suggests that explainable modelling is the preferred method.

Some argue explainable modelling is the only good choice in high-stakes domains [76]. However, prioritizing explainability at the cost of accuracy can also be argued to be unethical and some experiments show people prefer to prioritize the accuracy of the system in health care [78]. Although post-hoc explanations can be misleading, a potential solution could be to develop post-hoc explanation methods that include argumentative support for their claims [75]. If one wants to opt for a post-hoc explanation, our framework suggests that model-based explanations are the preferred type of explanation as they satisfy completeness and have quantitative proxy metrics available to evaluate soundness. Finally, we like to emphasize that the framework is a proposal that needs to be evaluated and possibly refined in practice.

More research is needed to investigate the performance of explainable models (e.g. rule-based systems, generalized additive models with or without interaction terms) in the health-care domain [7] and to improve explainable modelling methods. Furthermore, as interpretable features lead to more interpretable explanations, interpretable feature engineering is also important. We believe developing hybrid methods with data-driven and knowledge-driven elements for feature selection or engineering is another promising research direction to enhance interpretability.

When using explainable AI to create trustworthy AI, evaluating the quality of explanations is key. This part of the literature is currently underdeveloped and there are no standard evaluation methods yet in the community [14, 16]. We outlined several properties of explainability that are important (Section 2), and assessed the current state of quantitative evaluation metrics (Section 5). It should be noted that although quantitative proxy metrics are necessary for an objective assessment of explanation quality and a formal comparison of explanation methods, they should be complemented with human evaluation methods before employing AI systems in real-life. Good performance may not give direct evidence of success [79*], for example, Poursabzi-Sangdeh et al. [28*] find that explanations do not necessarily lead to better human-machine task performance. We found that clarity is difficult to assess for local explanations and quantitative evaluation metrics for example-based methods are still lacking. Another problem when evaluating the quality of explanations is that although interpretability is generally accepted to be user dependent, it is not quantified as such [14]. Determining a standard that indicates when AI systems are explainable for different users is thus an important direction for future work. Furthermore, outlining how different explanations can be best combined in a user interface [80], and how these combined AI systems should be evaluated are open research problems.

### 7.2 Complementary measures to create trustworthy AI in health care

Some argue that explanations are neither necessary, nor sufficient to establish trust in AI (e.g. [81*]). Other important influences at play are perceived system ability, control, and predictability [82]. Hence, although the field of explainable AI can contribute to trustworthy AI, it has its limits [83]. In this section, we highlight some other measures that can be used complementary to create trustworthy AI in health care:

- Reporting data quality. As real-world data are not collected for research purposes they may contain biases, mistakes, or be incomplete. Hence, understanding the data quality and how the data were collected is at least as important as explainability since it allows to understand the limitations of the resulting model [78]. The same dataset can be high quality for one purpose and low quality for another. A widely accepted framework that can be used to assess whether EHR data is suited for a specific use case is presented by Kahn et al. [84]. This framework evaluates data quality based on conformance, completeness, and plausibility, and can be used to communicate the findings in a structured manner to users of the AI system that uses the data.

- Performing extensive (external) validation. The concern that models are not robust or generalizable, can be addressed using external validation. Replicating a prediction model on new data can be a slow process due to lack of data standardization. Although external validation is recognized as an area of improvement for clinical risk prediction models [85], it is increasingly feasible with the adoption of common data structures in health care (e.g. OMOP-CDM [86]). The Observational Health Data Sciences and Informatics (OHDSI) network has developed standards and tools that allow patient-level prediction models to be developed and externally validated at a large scale in a transparent way following accepted best practices [87, 88]. This also ensures reproducibility of the results. Other model desiderata can likewise be assessed directly instead of by using explainability. Research in this area is for example addressing stability [89], fairness [90], and privacy [91]. Developing quality checks for these model desiderata, as well as investigating how to incorporate model desiderata during model optimization, are promising alternatives to create more trustworthy AI.

- Regulation. Although regulation of AI systems is currently still under development, established regulation for other safety-critical applications (e.g. drug safety) suggests that it can be an effective way to generate trust in the long-term. There are different possible forms of regulating AI. The first way is requiring the AI system to satisfy pre-defined requirements. However, it is difficult to define an exhaustive list of verifiable criteria that ensure an AI system is lawful, ethical, and robust. Instead of regulating the end-product (i.e. the AI system), an alternative would be to control the development process by introducing standard development guidelines that should be followed. However, just like it is difficult to assess model quality on all desired points, it might be difficult to get sufficient insight in the development process. Finally, we could introduce a licensing system to regulate developers as suggested by Mittelstadt [9]. This allows professional accountability (e.g. malpractice can be punished with losing one's license) and can be compared to licensing of doctors in the health-care domain. AI systems used in health care are currently regulated as medical devices, but this regulation needs to be adjusted to suit the adaptive nature of AI systems [92]. The U.S. Food and Drug Administration is currently investigating new types of regulation for digital technologies, among which a shift of regulation from end-products to firms (developers) [93].

## 8. Conclusion

The aim of this paper is to provide guidance to researchers and practitioners on the design of explainable AI systems for the health-care domain and contribute to formalization of the field of explainable AI. This survey provides a holistic view of the literature; connecting different perspectives and providing concrete design recommendations. We conclude that explainable modelling might be preferred over post-hoc explanations when using explainable AI to create trustworthy AI for health care. In addition, we find that evidence of the usefulness of explainability is still lacking in practice and recognize that complementary measures might be needed to create trustworthy AI (e.g. reporting data quality, performing extensive (external) validation, and regulation).

## Acknowledgements
The authors like to thank Dr. Jenna Reps for her valuable feedback on this manuscript.

## Declaration of competing interest
The authors declare that they have no known competing financial interests or personal relationships that could have appeared to influence the work reported in this paper.

## Funding
This project has received support from the European Health Data and Evidence Network (EHDEN) project. EHDEN received funding from the Innovative Medicines Initiative 2 Joint Undertaking (JU) under grant agreement No 806968. The JU receives support from the European Union's Horizon 2020 research and innovation programme and EFPIA.

## References
[1] International Data Corporation. Worldwide spending on artificial intelligence systems will be nearly $98 billion in 2023, according to new IDC spending guide (2019). Accessed: July 4, 2020. https://www.idc.com/getdoc.jsp?containerId=prUS45481219.

[2] Rajkomar A, Oren E, Chen K, Dai AM, Hajaj N, Hardt M, et al. Scalable and accurate deep learning with electronic health records, NPJ Digit. Med. 1(2018) pp. 1-18. https://doi.org/10.1038/s41746-018-0029-1.

[3] Tonekaboni S, Joshi S, McCradden MD, Goldenberg A. What clinicians want: Contextualizing explainable machine learning for clinical end use, Proceedings of Machine Learning research. (2019) pp. 1-21.

[4] Peterson ED. Machine learning, predictive analytics, and clinical practice: Can the past inform the present?, JAMA. 322(2019) pp. 2283-2284. https://doi.org/10.1001/jama.2019.17831.

[5] He J, Baxter SL, Xu J, Xu J, Zhou X, Zhang K. The practical implementation of artificial intelligence technologies in medicine, Nat. Med. 25(2019) pp. 30-36. https://doi.org/10.1038/s41591-018-0307-0.

[6] Topol EJ. High-performance medicine: The convergence of human and artificial intelligence, Nat. Med. 25(2019) pp. 44-56. https://doi.org/10.1038/s41591-018-0300-7.

[7] Ahmad MA, Eckert C, Teredesai A. Interpretable machine learning in healthcare, Proceedings of the 2018 ACM International Conference on Bioinformatics, Computational Biology, and Health Informatics. (2018) pp. 559-560. https://doi.org/10.1145/3233547.3233667.

[8] European Commission. High level expert group on artificial intelligence. Ethics guidelines for trustworthy AI. Published in Brussels: April 8, 2019. https://ec.europa.eu/newsroom/dae/document.cfm?doc_id=60419.

[9] Mittelstadt B. Principles alone cannot guarantee ethical AI, Nat. Mach. Intell. (2019) pp. 501-507. https://doi.org/10.1038/s42256-019-0114-4.

[10] European Institute of Innovation and Technology Health. Transforming healthcare with AI: The impact on the workforce and organisations. Published: March, 2020. https://eithealth.eu/wp-content/uploads/2020/03/EIT-Health-and-McKinsey_Transforming-Healthcare-with-AI.pdf.

[11] Select Committee on Artificial Intelligence. The national artificial intelligence research and development strategic plan: 2019 update. Published: June, 2019. https://www.nitrd.gov/pubs/National-AI-RD-Strategy-2019.pdf.

[12] Lipton ZC. The mythos of model interpretability, Queue. (2018) pp. 31–57. https://doi.org/10.1145/3236386.3241340.

[13] Doshi-Velez F, Kim B. Towards a rigorous science of interpretable machine learning [Preprint]. March 2, 2017. https://arxiv.org/abs/1702.08608

[14] Guidotti R, Monreale A, Ruggieri S, Turini F, Giannotti F, Pedreschi D. A survey of methods for explaining black box models, ACM computing surveys (CSUR). 51(2018) pp. 93. https://doi.org/10.1145/3236009.

[15] Mohseni S, Zarei N, Ragan ED. A survey of evaluation methods and measures for interpretable machine learning [Preprint]. April 26, 2020. https://arxiv.org/pdf/1811.11839v4.pdf

[16] Adadi A, Berrada M. Peeking inside the black-box: A survey on explainable artificial intelligence (XAI), IEEE Access. 6(2018) pp. 52138-52160. https://doi.org/10.1109/access.2018.2870052.

[17] Arrieta AB, Díaz-Rodríguez N, Del Ser J, Bennetot A, Tabik S, Barbado A, et al. Explainable artificial intelligence (XAI): Concepts, taxonomies, opportunities and challenges toward responsible AI, Inf. Fusion. 58(2019) pp. 82-115. https://doi.org/10.1016/j.inffus.2019.12.012.

[18] Carvalho DV, Pereira EM, Cardoso JS. Machine learning interpretability: A survey on methods and metrics, Electronics. 8(2019) pp. 832. https://doi.org/10.3390/electronics8080832.

[19] Murdoch WJ, Singh C, Kumbier K, Abbasi-Asl R, Yu B. Definitions, methods, and applications in interpretable machine learning, Proceedings of the National Academy of Sciences. 116(2019) pp. 22071-22080. https://doi.org/10.1073/pnas.1900654116.

[20] Payrovnaziri SN, Chen Z, Rengifo-Moreno P, Miller T, Bian J, Chen JH, et al. Explainable artificial intelligence models using real-world electronic health record data: A systematic scoping review, J. Am. Med. Inform. Assoc. (2020). https://doi.org/10.1093/jamia/ocaa053.

[21] Liberati A, Altman DG, Tetzlaff J, Mulrow C, Gøtzsche PC, Ioannidis JP, et al. The PRISMA statement for reporting systematic reviews and meta-analyses of studies that evaluate health care interventions: Explanation and elaboration, J. Clin. Epidemiol. 62(2009) pp. e1-e34, doi:10.1016/j.jclinepi.2009.06.006.

[22] Doran D, Schulz S, Besold TR. What does explainable AI really mean? A new conceptualization of perspectives [Preprint]. October 2, 2017. https://arxiv.org/abs/1710.00794

[23] Cabitza F, Campagner A, Ciucci D. New frontiers in explainable AI: Understanding the gi to interpret the go, International Cross-Domain Conference for Machine Learning and Knowledge Extraction. (2019) pp. 27-47. https://doi.org/10.1007/978-3-030-29726-8_3.

[24] Miller T. Explanation in artificial intelligence: Insights from the social sciences, Artificial Intelligence. 267(2019) pp. 1-38. https://doi.org/10.1016/j.artint.2018.07.007.

[25] Gilpin LH, Bau D, Yuan BZ, Bajwa A, Specter M, Kagal L. Explaining explanations: An approach to evaluating interpretability of machine learning, 2018 IEEE 5th International Conference on Data Science and Advanced Analytics (DSAA). (2018). https://doi.org/10.1109/DSAA.2018.00018.

[26] Ribeiro MT, Singh S, Guestrin C. Why should i trust you?: Explaining the predictions of any classifier, Proceedings of the 22nd ACM SIGKDD International Conference on Knowledge Discovery and Data Mining. (2016) pp. 1135-1144. https://doi.org/10.18653/v1/N16-3020

[27] Mueller ST, Hoffman RR, Clancey W, Emrey A, Klein G. Explanation in human-AI systems: A literature meta-review, synopsis of key ideas and publications, and bibliography for explainable AI [Preprint]. Februrary 9, 2019. https://arxiv.org/pdf/1902.01876.pdf

[28] Poursabzi-Sangdeh F, Goldstein DG, Hofman JM, Vaughan JW, Wallach H. Manipulating and measuring model interpretability [Preprint]. November 8, 2018. https://arxiv.org/pdf/1802.07810.pdf

[29] Ras G, van Gerven M, Haselager P. Explanation methods in deep learning: Users, values, concerns and challenges, in: Explainable and interpretable models in computer vision and machine learning: Springer, 2018. p. 19-36.

[30] Kulesza T, Stumpf S, Burnett M, Yang S, Kwan I, Wong W-K. Too much, too little, or just right? Ways explanations impact end users' mental models, 2013 IEEE Symposium on Visual Languages and Human Centric Computing. (2013) pp. 3-10. https://doi.org/10.1109/VLHCC.2013.6645235.

[31] Weld DS, Bansal G. The challenge of crafting intelligible intelligence [Preprint]. October 15, 2018. https://arxiv.org/abs/1803.04263

[32] Lou Y, Caruana R, Gehrke J, Hooker G. Accurate intelligible models with pairwise interactions, Proceedings of the 19th ACM SIGKDD international conference on Knowledge discovery and data mining. (2013) pp. 623-631. https://doi.org/10.1145/2487575.2487579.

[33] Samek W. Explainable AI: Interpreting, explaining and visualizing deep learning: Springer Nature; 2019.

[34] Goodman B, Flaxman S. European Union regulations on algorithmic decision-making and a "right to explanation", AI Magazine. 38(2017) pp. 50-57. https://doi.org/10.1609/aimag.v38i3.2741.

[35] Bunt A, Lount M, Lauzon C. Are explanations always important?: A study of deployed, low-cost intelligent interactive systems, Proceedings of the 2012 ACM International Conference on Intelligent User Interfaces. (2012) pp. 169-178. https://doi.org/10.1145/2166966.2166996.

[36] Tjoa E, Guan C. A survey on explainable artificial intelligence (XAI): Towards medical XAI [Preprint]. June 7, 2020. https://arxiv.org/pdf/1907.07374.pdf

[37] Lakkaraju H, Kamar E, Caruana R, Leskovec J. Interpretable & explorable approximations of black box models [Preprint]. July 4, 2017. https://arxiv.org/pdf/1707.01154.pdf

[38] Bastani O, Kim C, Bastani H. Interpreting blackbox models via model extraction [Preprint]. January 24, 2017. https://arxiv.org/abs/1904.11829

[39] Tan S, Caruana R, Hooker G, Lou Y. Distill-and-compare: Auditing black-box models using transparent model distillation, Proceedings of the 2018 AAAI/ACM Conference on AI, Ethics, and Society. (2018) pp. 303-310.

[40] Alaa AM, van der Schaar M. Demystifying black-box models with symbolic metamodels, Advances in Neural Information Processing Systems 33. (2019) pp. 11304-11314.

[41] Ribeiro MT, Singh S, Guestrin C. Anchors: High-precision model-agnostic explanations, Thirty-Second AAAI Conference on Artificial Intelligence. (2018).

[42] Friedman JH. Greedy function approximation: A gradient boosting machine, Ann. Stat. (2001) pp. 1189-1232. https://doi.org/10.1214/aos/1013203451.

[43] Friedman JH, Popescu BE. Predictive learning via rule ensembles, Ann. Appl. Stat. 2(2008) pp. 916-954. https://doi.org/10.1214/07-AOAS148.

[44] Apley DW, Zhu J. Visualizing the effects of predictor variables in black box supervised learning models [Preprint]. August 18, 2019. https://arxiv.org/pdf/1612.08468.pdf

[45] Fisher A, Rudin C, Dominici F. All models are wrong but many are useful: Variable importance for black-box, proprietary, or misspecified prediction models, using model class reliance [Preprint]. December 23, 2018. https://arxiv.org/pdf/1801.01489.pdf

[46] Lei J, G'Sell M, Rinaldo A, Tibshirani RJ, Wasserman L. Distribution-free predictive inference for regression, J. Am. Stat. Assoc. 113(2018) pp. 1094-1111. https://doi.org/10.1080/01621459.2017.1307116.

[47] Goldstein A, Kapelner A, Bleich J, Pitkin E. Peeking inside the black box: Visualizing statistical learning with plots of individual conditional expectation, J. Comput. Graph. Stat. 24(2015) pp. 44-65. https://doi.org/10.1080/10618600.2014.907095.

[48] Datta A, Sen S, Zick Y. Algorithmic transparency via quantitative input influence: Theory and experiments with learning systems, 2016 IEEE Symposium on Security and Privacy. (2016) pp. 598-617. https://doi.org/10.1007/978-3-319-54024-5_4.

[49] Lundberg SM, Lee S-I. A unified approach to interpreting model predictions, Advances in Neural Information Processing Systems 30. (2017) pp. 4765-4774.

[50] Yoon J, Jordon J, van der Schaar M. Invase: Instance-wise variable selection using neural networks, International Conference of Learning Representations. (2018) pp. 1-24.

[51] Cook RD. Detection of influential observation in linear regression, Technometrics. 19(1977) pp. 15-18. https://doi.org/10.1080/00401706.1977.10489493.

[52] Kim B, Khanna R, Koyejo OO. Examples are not enough, learn to criticize! Criticism for interpretability, Advances in Neural Information Processing Systems 29. (2016) pp. 2280-2288.

[53] Wachter S, Mittelstadt B, Russell C. Counterfactual explanations without opening the black box: Automated decisions and the gpdr, Harv. J. L. & Tech. 31(2017) pp. 841-887. https://doi.org/10.2139/ssrn.3063289.

[54] Kim B. Interactive and interpretable machine learning models for human machine collaboration [Doctoral dissertation]: Massachusetts Institute of Technology, 2015.

[55] Caruana R, Lou Y, Gehrke J, Koch P, Sturm M, Elhadad N. Intelligible models for healthcare: Predicting pneumonia risk and hospital 30-day readmission, Proceedings of the 21th ACM SIGKDD International Conference on Knowledge Discovery and Data Mining. (2015) pp. 1721-1730. https://doi.org/10.1145/2783258.2788613.

[56] Wu M, Hughes MC, Parbhoo S, Zazzi M, Roth V, Doshi-Velez F. Beyond sparsity: Tree regularization of deep models for interpretability, Thirty-Second AAAI Conference on Artificial Intelligence. (2018).

[57] Zhang Q, Nian Wu Y, Zhu S-C. Interpretable convolutional neural networks, Proceedings of the IEEE Conference on Computer Vision and Pattern Recognition. (2018) pp. 8827-8836. https://doi.org/10.1109/CVPR.2018.00920.

[58] Vaughan J, Sudjianto A, Brahimi E, Chen J, Nair VN. Explainable neural networks based on additive index models [Preprint]. June 5, 2018. https://arxiv.org/pdf/1806.01933.pdf

[59] Wang T. Hybrid decision making: When interpretable models collaborate with black-box models, J. Mach. Learn. Res. (2019) pp. 1-31.

[60] Hind M, Wei D, Campbell M, Codella NC, Dhurandhar A, Mojsilović A, et al. Ted: Teaching AI to explain its decisions, Proceedings of the 2019 AAAI/ACM Conference on AI, Ethics, and Society. (2019) pp. 123-129. https://doi.org/10.1145/3306618.3314273.

[61] Che Z, Purushotham S, Khemani R, Liu Y. Interpretable deep models for ICU outcome prediction, AMIA Annual Symposium Proceedings. (2016) pp. 371-380.

[62] Ancona M, Ceolini E, Öztireli C, Gross M. Towards better understanding of gradient-based attribution methods for deep neural networks [Preprint]. March 7, 2018. https://arxiv.org/abs/1711.06104

[63] Pan L, Liu G, Mao X, Li H, Zhang J, Liang H, et al. Development of prediction models using machine learning algorithms for girls with suspected central precocious puberty: Retrospective study, JMIR Med. Inform. 7(2019) pp. 1-13. https://doi.org/10.2196/11728.

[64] Katuwal GJ, Chen R. Machine learning model interpretability for precision medicine [Preprint]. October 28, 2016. https://arxiv.org/pdf/1610.09045.pdf

[65] Ghafouri-Fard S, Taheri M, Omrani MD, Daaee A, Mohammad-Rahimi H, Kazazi H. Application of single-nucleotide polymorphisms in the diagnosis of autism spectrum disorders: A preliminary study with artificial neural networks, J. Mol. Neurosci. 68(2019) pp. 515-521. https://doi.org/10.1007/s12031-019-01311-1.

[66] Huysmans J, Dejaeger K, Mues C, Vanthienen J, Baesens B. An empirical evaluation of the comprehensibility of decision table, tree and rule based predictive models, Decis. Support Syst. 51(2011) pp. 141-154. https://doi.org/10.1016/j.dss.2010.12.003.

[67] Lage I, Chen E, He J, Narayanan M, Kim B, Gershman S, et al. An evaluation of the human-interpretability of explanation [Preprint]. August 28, 2019. https://arxiv.org/pdf/1902.00006.pdf

[68] Friedler SA, Roy CD, Scheidegger C, Slack D. Assessing the local interpretability of machine learning models [Preprint]. August 2, 2019. https://arxiv.org/abs/1902.03501

[69] Molnar C, Casalicchio G, Bischl B. Quantifying interpretability of arbitrary machine learning models through functional decomposition [Preprint]. September 23, 2019. https://arxiv.org/pdf/1904.03867.pdf

[70] Arras L, Osman A, Müller K-R, Samek W. Evaluating recurrent neural network explanations [Preprint]. June 4, 2019. https://arxiv.org/abs/1904.11829

[71] Montavon G, Lapuschkin S, Binder A, Samek W, Müller K-R. Explaining nonlinear classification decisions with deep taylor decomposition, Pattern Recognit. 65(2017) pp. 211-222. https://doi.org/10.1016/j.patcog.2016.11.008.

[72] Hooker S, Erhan D, Kindermans P-J, Kim B. A benchmark for interpretability methods in deep neural networks, Advances in Neural Information Processing Systems 33. (2018).

[73] Sundararajan M, Taly A, Yan Q. Axiomatic attribution for deep networks, Proceedings of the 34th International Conference on Machine Learning. (2017) pp. 3319-3328.

[74] Montavon G, Samek W, Müller K-R. Methods for interpreting and understanding deep neural networks, Digit. Signal Process. 73(2018) pp. 1-15. https://doi.org/10.1016/j.dsp.2017.10.011.

[75] Mittelstadt B, Russell C, Wachter S. Explaining explanations in AI, Proceedings of the Conference on Fairness, Accountability, and Transparency. (2019) pp. 279-288. https://doi.org/10.1145/3287560.3287574.

[76] Rudin C. Stop explaining black box machine learning models for high stakes decisions and use interpretable models instead, Nat. Mach. Intell. 1(2019) pp. 206-215. https://doi.org/10.1038/s42256-019-0048-x.

[77] European Commission. White paper: On artificial intelligence - a European approach to excellence and trust. Published in Brussels: February 19, 2020. https://ec.europa.eu/info/sites/info/files/commission-white-paper-artificial-intelligence-feb2020_en.pdf.

[78] The Royal Society. Explainable AI: The basics. Published: November, 2019. https://royalsociety.org/-/media/policy/projects/explainable-ai/AI-and-interpretability-policy-briefing.pdf.

[79] Buçinca Z, Lin P, Gajos KZ, Glassman EL. Proxy tasks and subjective measures can be misleading in evaluating explainable AI systems [Preprint]. January 22, 2020. https://arxiv.org/pdf/2001.08298.pdf

[80] Hohman F, Head A, Caruana R, DeLine R, Drucker SM. Gamut: A design probe to understand how data scientists understand machine learning models, Proceedings of the 2019 CHI Conference on Human Factors in Computing Systems. (2019) pp. 1-13. https://doi.org/10.1145/3290605.3300809.

[81] Hall P, Gill N, Schmidt N. Proposed guidelines for the responsible use of explainable machine learning [Preprint]. November 29, 2019. https://arxiv.org/pdf/1906.03533.pdf

[82] Holliday D, Wilson S, Stumpf S. User trust in intelligent systems: A journey over time, Proceedings of the 21st International Conference on Intelligent User Interfaces. (2016) pp. 164-168. https://doi.org/10.1145/2856767.2856811.

[83] Sendak M, Elish MC, Gao M, Futoma J, Ratliff W, Nichols M, et al. "The human body is a black box" supporting clinical decision-making with deep learning, Proceedings of the 2020 Conference on Fairness, Accountability, and Transparency. (2020) pp. 99-109. https://doi.org/10.1145/3351095.3372827.

[84] Kahn MG, Callahan TJ, Barnard J, Bauck AE, Brown J, Davidson BN, et al. A harmonized data quality assessment terminology and framework for the secondary use of electronic health record data, EGEMS. 4(2016) pp. 1244. https://doi.org/10.13063/2327-9214.1244.

[85] Goldstein BA, Navar AM, Pencina MJ, Ioannidis J. Opportunities and challenges in developing risk prediction models with electronic health records data: A systematic review, J. Am. Med. Inform. Assoc. 24(2017) pp. 198-208. https://doi.org/10.1093/jamia/ocw042.

[86] Overhage JM, Ryan PB, Reich CG, Hartzema AG, Stang PE. Validation of a common data model for active safety surveillance research, J. Am. Med. Inform. Assoc. 19(2012) pp. 54-60. https://doi.org/10.1136/amiajnl-2011-000376.

[87] Reps JM, Schuemie MJ, Suchard MA, Ryan PB, Rijnbeek PR. Design and implementation of a standardized framework to generate and evaluate patient-level prediction models using observational healthcare data, J. Am. Med. Inform. Assoc. 25(2018) pp. 969-975. https://doi.org/10.1093/jamia/ocy032.

[88] Reps JM, Williams RD, You SC, Falconer T, Minty E, Callahan A, et al. Feasibility and evaluation of a large-scale external validation approach for patient-level prediction in an international data network: Validation of models predicting stroke in female patients newly diagnosed with atrial fibrillation, BMC Med. Res. Methodol. 20(2020) pp. 102-102. https://doi.org/10.1186/s12874-020-00991-3.

[89] Philipp M, Rusch T, Hornik K, Strobl C. Measuring the stability of results from supervised statistical learning, J. Comput. Graph. Stat. 27(2018) pp. 685-700. https://doi.org/10.1080/10618600.2018.1473779.

[90] Hardt M, Price E, Srebro N. Equality of opportunity in supervised learning, Proceedings of the 30th International Conference on Neural Information Processing Systems. (2016) pp. 3315-3323.

[91] Dwork C, Hardt M, Pitassi T, Reingold O, Zemel R. Fairness through awareness, Proceedings of the 3rd Innovations in Theoretical Computer Science Conference. (2012) pp. 214-226. https://doi.org/10.1145/2090236.2090255.

[92] US Food and Drug Administration. Proposed regulatory framework for modifications to artificial intelligence/machine learning (AI/ML) - based software as a medical device (SAMD). Published: January, 2020. https://www.fda.gov/media/122535/download.

[93] Cortez N. Digital health and regulatory experimentation at the FDA, Yale Journal of Law & Technology. 21(2019).